\documentclass[10pt,twocolumn,letterpaper]{article}

\usepackage{wacv}      

\usepackage{listings}
\usepackage{xcolor}
\usepackage{booktabs}
\usepackage{multirow}
\usepackage{longtable}
\usepackage{tabularx}
\usepackage[table]{xcolor} 
\usepackage[accsupp]{axessibility}  

\definecolor{wacvblue}{rgb}{0.21,0.49,0.74}
\usepackage[pagebackref,breaklinks,colorlinks,allcolors=wacvblue]{hyperref}

\usepackage[most]{tcolorbox}

\def\wacvPaperID{2885} 
\def\confName{WACV}
\def\confYear{2026}

\title{Exploring Automated Recognition of Instructional Activity\\and Discourse from Multimodal Classroom Data}

\author{
  \textbf{Ivo Bueno\textsuperscript{1,2}\thanks{These authors contributed equally to this work.}}~
  \textbf{Ruikun Hou\textsuperscript{1,2}\footnotemark[1]}~
  \textbf{Babette Bühler\textsuperscript{1}}~
  \textbf{Tim Fütterer\textsuperscript{3}}
  \textbf{James Drimalla\textsuperscript{4}}~
\\
  \textbf{Jonathan K. Foster\textsuperscript{5}}~
  \textbf{Peter Youngs\textsuperscript{6}}~
  \textbf{Peter Gerjets\textsuperscript{7}}~
  \textbf{Ulrich Trautwein\textsuperscript{3}}~
  \textbf{Enkelejda Kasneci\textsuperscript{1,2}}
\\
  \textsuperscript{1}Technical University of Munich~
  \textsuperscript{2}Munich Center for Machine Learning (MCML)~ 
\\
  \textsuperscript{3}University of Tübingen~
  \textsuperscript{4}Gordon College~
  \textsuperscript{5}University at Albany~
\\
  \textsuperscript{6}University of Virginia~
  \textsuperscript{7}Leibniz-Institut für Wissensmedien~
\\
  \small{
    \href{mailto:ivo.bueno@tum.de}{ivo.bueno@tum.de},
    \href{mailto:ruikun.hou@tum.de}{ruikun.hou@tum.de}
  }
}

\begin{document}
\maketitle
\begin{abstract}
Observation of classroom interactions can provide concrete feedback to teachers, but current methods rely on manual annotation, which is resource-intensive and hard to scale. This work explores AI-driven analysis of classroom recordings, focusing on multimodal instructional activity and discourse recognition as a foundation for actionable feedback. Using a densely annotated dataset of 164 hours of video and 68 lesson transcripts, we design parallel, modality-specific pipelines. For video, we evaluate zero-shot multimodal LLMs, fine-tuned vision–language models, and self-supervised video transformers on 24 activity labels. For transcripts, we fine-tune a transformer-based classifier with contextualized inputs and compare it against prompting-based LLMs on 19 discourse labels. To handle class imbalance and multi-label complexity, we apply per-label thresholding, context windows, and imbalance-aware loss functions. The results show that fine-tuned models consistently outperform prompting-based approaches, achieving macro-F1 scores of 0.577 for video and 0.460 for transcripts. These results demonstrate the feasibility of automated classroom analysis and establish a foundation for scalable teacher feedback systems.
\end{abstract}
    
\section{Introduction}
\label{sec:intro}
High-quality feedback to teachers on their instructional practices is essential for improving teaching quality and is strongly associated with enhanced student achievement~\cite{kraft2018theeffectofteachercoaching, steinberg2015doesteacherevaluation}. However, current approaches to teacher feedback rely heavily on manual observation by trained annotators, either in real time or through reviewing classroom video recordings~\cite{goldberg2021attentive,sumer2021multimodal,buhler2023automated}. Such methods are resource-intensive and impractical to scale to the frequency and granularity required for effective professional development~\cite{ramakrishnan2021toward,hou2024automatedassessment,futterer2026validating}. 

Recent advances in artificial intelligence (AI), particularly in computer vision (CV) and natural language processing (NLP), have transformed high-stakes domains such as autonomous driving and medical diagnosis. This has sparked growing interest in leveraging these techniques in education to support and enhance teaching and learning~\cite{kasneci2023chatgpt,futterer2025artificial,sessler2025can,sumer2018teachers,buhler2025lab,bozkir2025automated}. In this work, we investigate how state-of-the-art AI models can analyze multimodal classroom recordings—videos and transcripts—to achieve fine-grained recognition of instructional practices.

Crucially, we emphasize that the automatic recognition of teaching practices does not by itself constitute actionable feedback. Instead, it serves as scalable groundwork for systematic assessment by enabling the capture of classroom activities and discourse features at a scale infeasible through manual observation~\cite{hou2025multimodalassessmentclassroomdiscourse,futterer2026validating}. This capability not only lays the foundation for higher-level interpretation, such as evaluating teaching quality and designing evidence-based feedback strategies, but also opens new opportunities for large-scale research on instructional practices across diverse educational contexts \cite{foster2024automated,demszky2024can,whitehill2023automated}. 
Translating this vision into practice requires addressing a concrete technical task: the accurate recognition of instructional practices from classroom recordings. Real-world classroom interactions consist of both visual activities (e.g., student engagement, teacher movement) and linguistic discourse (e.g., questioning strategies, feedback moves) \cite{wang2024artificial,hou2025multimodalassessmentclassroomdiscourse}. Identifying these multimodal signals reliably and at scale is therefore a prerequisite for developing feedback systems that can generalize across classrooms and for advancing empirical research on teaching quality. This recognition task forms the central problem we address in this work.

Classroom data analysis presents several unique challenges. First, both activity and discourse recognition are inherently multi-label tasks, with each instance potentially involving multiple overlapping instructional practices. Second, the label distributions are highly imbalanced, with negative instances far outnumbering positives within each label, further complicating learning. Third, both modalities provide limited context: vision models classify single one-second clips, and transcript models operate on sentence-level discourse units, making it difficult to capture broader pedagogical structures. Fourth, classrooms are visually complex environments, with variable camera angles, frequent occlusions caused by teacher and student movement, and additional challenges in recognizing children, whose gestures and expressions are less standardized than those of adults \cite{sumer2021multimodal,goldberg2021attentive}. Variability in classroom layouts, lighting conditions, and instructional materials (e.g., whiteboards, projectors, manipulatives) further complicates robust recognition \cite{drimalla2025investigating}. Finally, recordings come from diverse classrooms and teaching conditions, introducing domain shifts that affect both visual and textual modalities and require models to generalize across heterogeneous contexts.

\begin{figure*}[t]
    \centering
    \includegraphics[width=0.75\linewidth, trim=1.5cm 5cm 1cm 1.5cm, clip]{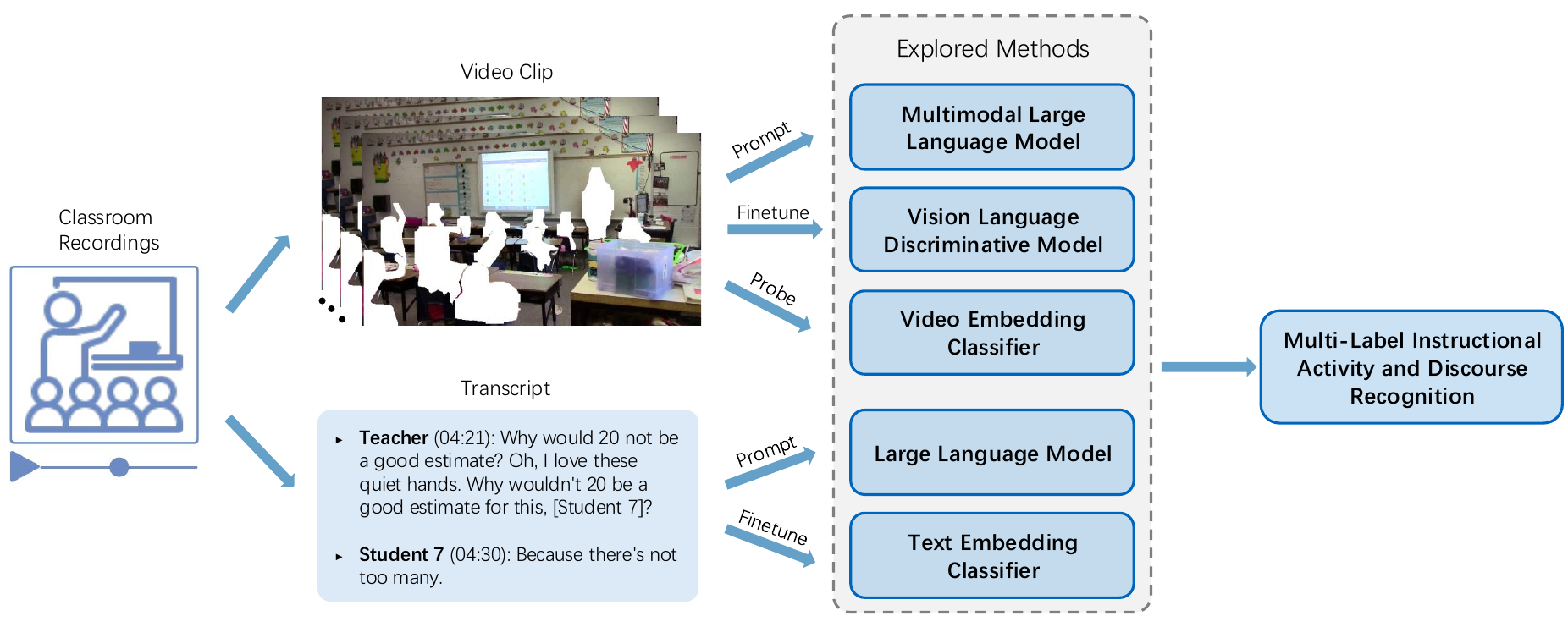}
    \caption{Overview of explored approaches to multi-label activity and discourse recognition in classroom video and transcripts. Individuals are removed for privacy.}
    \label{fig:pipeline}
\end{figure*}

To address these challenges, we adopt a parallel, modality-specific pipeline rather than an early-fusion strategy. Figure~\ref{fig:pipeline} contains a visual representation of this pipeline. Each modality is optimized independently with architectures and training strategies tailored to its properties. For vision, we implement advanced video recognition approaches, including zero-shot prompting of multimodal large language models (MLLMs), fine-tuning of vision–language discriminative models, and attentive probing of pre-trained video transformer encoders for instructional activity recognition. For transcripts, we employ zero- and few-shot prompting of LLMs and fine-tune an embedding-based classifier. We incorporate short visual temporal windows and prepend the two previous utterances as contextual input to enrich both video and textual representations. Finally, we perform per-label dynamic thresholding during evaluation, selecting thresholds that maximize the macro-F1 score and adapting to label-specific imbalances.

Our contributions are threefold:
\begin{enumerate}
    \item We adapt state-of-the-art vision and language models to the domain of elementary school math and English language arts (ELA) lessons through parallel, modality-specific pipelines.  
    \item We provide a systematic comparison of prompting-based LLMs and fine-tuned specialized architectures for video and transcript data, providing insights into their relative strengths and limitations in this educational setting.  
    \item We evaluate practical strategies for addressing multi-label imbalance, including per-label thresholding, context windows, and imbalance-aware loss functions, offering practical guidance for future applications in classroom discourse and activity recognition.  
\end{enumerate}


\section{Related Work}
\label{sec:relatedwork}
\subsection{Video Action Recognition}
Action recognition has been a fundamental CV task, aiming to classify human actions and events in videos. Effective exploitation of both spatial and temporal information is essential for constructing robust video recognition models. 
Early research relied on 2D-CNNs combined with recurrent architectures such as LSTMs to capture temporal dependencies across frames~\cite{gammulle2017two,wu2015modeling}. 
Subsequently, 3D-CNNs were introduced to jointly model spatio-temporal features~\cite{feichtenhofer2019slowfast,carreira2017quo}. 
Given the limited receptive fields of CNN-based architectures, transformer models have recently become dominant in visual understanding. 
Vision transformers (ViTs)~\cite{dosovitskiy2020image} partitioned images into a set of patches, embedded them as tokens, and applied self-attention~\cite{vaswani2017attention} to capture long-range relationships.
Their extensions to the video domain, such as \textit{ViViT}~\cite{arnab2021vivit} and \textit{TimeSformer}~\cite{bertasius2021space}, leveraged temporal attention mechanisms and showed promising performance on standard video benchmarks like \textit{Kinetics-400}~\cite{carreira2017quo} and \textit{Something-Something v2 (SSv2)}~\cite{goyal2017something}. 
More recently, self-supervised video representation learning like \textit{VideoMAE}~\cite{tong2022videomae} and \textit{V-JEPA}~\cite{bardes2024revisiting} enabled scalable pretraining across massive video corpora, achieving strong generalizable performance across downstream recognition tasks. 

In parallel, vision-language models (VLMs) have grown increasingly influential. Contrastive language-image pre-training (CLIP)~\cite{radford2021learning} demonstrated the effectiveness of learning joint embedding spaces from large-scale image-text pairs. Several extensions adapted \textit{CLIP} to video analysis, including \textit{ActionCLIP}~\cite{wang2021actionclip} and \textit{X-CLIP}~\cite{XCLIP}, enabling zero-shot recognition via video-to-text retrieval and achieving competitive performance when fine-tuned on domain-specific data. 
Meanwhile, MLLMs like \textit{GPT-4V}~\cite{achiam2023gpt}, which combined vision encoders with LLMs, further expanded the possibilities of zero-shot video understanding through flexible prompting and natural language generation.

These advances in visual understanding have drawn growing attention to their applications in classroom scenarios, such as identifying teacher postures (e.g., writing on the blackboard and moving across the room) and student behaviors (e.g., hand-raising and taking notes). Prior research mainly focused on individual analysis by detecting each participant and then classifying their behaviors, applying various approaches ranging from CNN-based detectors~\cite{chen2022teacher,sumer2021multimodal} and skeleton-driven action recognition~\cite{ahuja2019edusense,buhler2023automated} to transformer-based architectures~\cite{wang2025scb,zhao2023bitnet}. In contrast, our work shifts the focus to whole-scene analysis, formulating this as a multi-label activity recognition task over a temporal clip, following Foster et al.~\cite{foster2024automated}. This enables the capture of not only individual behaviors but also key interactions between teachers, students, and objects, which are essential for a comprehensive understanding of teaching quality. To this end, our present work explores and benchmarks state-of-the-art video recognition approaches on this real-world educational application, involving VLMs and video transformers.

\subsection{Transcript Discourse Recognition}
Discourse recognition has long been studied in NLP, particularly in the context of dialogue act classification. Early approaches relied on feature-based models using lexical, syntactic, and prosodic cues~\cite{stolcke-etal-2000-dialogue}, but were limited by handcrafted features and poor generalization across domains. Transformer-based models, beginning with \textit{BERT}~\cite{devlin-etal-2019-bert}, marked a turning point by producing context-aware embeddings that serve as strong backbones for downstream tasks. Successive variants further improved efficiency and accuracy: \textit{ALBERT}~\cite{lan2020albertlitebertselfsupervised} introduced parameter sharing and a Sentence Ordering objective, \textit{RoBERTa}~\cite{liu2019robertarobustlyoptimizedbert} leveraged larger corpora and dynamic masking, and \textit{DeBERTa}~\cite{he2021debertadecodingenhancedbertdisentangled} added disentangled attention. Most recently, \textit{DeBERTaV3}~\cite{he2023debertav3improvingdebertausing} advanced the state of the art with Replaced Token Detection and gradient-disentangled embedding sharing, improving both efficiency and representation quality.

LLMs have opened new directions by enabling zero- and few-shot discourse classification through prompt-based learning. Models such as \textit{GPT}~\cite{gpt} and \textit{LLaMA}~\cite{touvron2023llamaopenefficientfoundation} allow label semantics to be directly incorporated into prompts, proving especially useful in low-resource or domain-specific contexts. For example, Chen et al.~\cite{chen2022weaklysuperviseddataaugmentation} showed that mixing few-shot examples with prompt-augmented data can match or exceed state-of-the-art results using only 10\% of the training data. Yet, LLM-based methods face challenges: predictions can be inconsistent, inference is costly, and privacy concerns arise when working with sensitive data such as classroom transcripts. Moreover, they do not fully resolve structural issues in discourse recognition, including multi-label classification, severe class imbalance, and limited sentence-level context. These challenges motivate complementary approaches such as imbalance-aware objectives, e.g., Focal Loss~\cite{focalloss}, Asymmetric Loss (ASL)~\cite{benbaruch2020asymmetric}, and contextualization through preceding utterances.

In the educational domain, discourse recognition is vital for analyzing instructional practices, including teacher questioning strategies and feedback moves. Classroom transcripts, however, exacerbate existing challenges: labels are highly imbalanced, multiple discourse acts may co-occur, and domain shifts across classrooms are common. Recent studies have begun to address these issues. Hou et al.~\cite{hou2025multimodalassessmentclassroomdiscourse} proposed a multimodal model combining a \textit{BERT}-based encoder with visual and acoustic features to score discourse categories in the GTI protocol~\cite{bell2018annexA}, finding that textual input was most predictive. Demszky and Hill~\cite{demszky-hill-2023-ncte} built a dataset of elementary math classroom transcripts and fine-tuned a \textit{RoBERTa} model, achieving moderate to high accuracy across discourse categories. Long et al.~\cite{long2024evaluating} evaluated \textit{GPT-4}~\cite{achiam2023gpt} on middle school math and Chinese lessons, showing high inter-coder reliability with humans while highlighting the importance of context and the difficulty of subjective labels. Collectively, these studies underscore the potential of both fine-tuned transformers and LLMs for classroom discourse analysis, while emphasizing ongoing challenges in context modeling, subjectivity, and cross-domain generalization.
\section{Methods}
\label{sec:methods}


\subsection{Dataset}
\label{sec:data}
The dataset originated from the Developing Ambitious Instruction project~\cite{youngs2022development}, comprising approximately 1000 hours of video recordings from mathematics and ELA lessons in elementary schools. For this work, we utilized a subset of 164 hours of videos across 195 lessons. In addition, the dataset provided human-transcribed audio for 68 of the 195 lessons, segmented by timestamps and attributed to either teacher or student turns. 
The videos were densely annotated at one-second intervals using ELAN~\cite{wittenburg2006elan}, an open-source annotation tool, marking the occurrence of 24 instructional activities within each second, such as \textit{small group activity}, and \textit{presentation with technology}. Foster et al.~\cite{foster2024analyzingtheconsistency} reported an average pairwise raw agreement for the 24 video instructional activities of 0.976, although there was a considerable variability for the pairwise positive agreement.
In the transcripts, each discourse turn was annotated with 19 discourse labels (e.g., \textit{elaborated feedback}, and \textit{open-ended questions}), which we kept at an utterance level instead of aligning them to the activity labels' seconds level.
A complete description of the 24 activities and 19 discourse labels is provided in the Supplementary Material. 
Both the instructional activity and discourse labels exhibited highly imbalanced, long-tail distributions. Some practices occurred frequently while others are rare, e.g., the activity label \textit{Teacher writing} appeared in only $\sim$4\% of the data, whereas \textit{Teacher standing} occurred in over 50\% (see Supplementary Material for full distributions). 

\subsection{Problem Formulation}

Let a classroom video be represented as a sequence of one-second clips
$X^v = \{x_t^v\}_{t=1}^{T}$, where $T$ is the video duration in seconds. 
We define the instructional activity label space as $\mathcal{Y}^v = \{L_1, \ldots, L_{N^v}\}$, where $N^v$ is the number of activity labels. 
Since multiple activities may co-occur, the task is formulated as a multi-label classification problem. 
Each clip $x_t^v$ is associated with a ground-truth annotation 
$y_t^v = \{y_{t,i}^v\}_{i=1}^{N^v}$, where $y_{t,i}^v \in \{0,1\}$ indicates the presence of label $i \in \mathcal{Y}^v$ at time $t$. 
Likewise, the classroom discourse is represented as a sequence of transcribed talk turns 
$X^d = \{x_k^d\}_{k=1}^{K}$, where each $x_k^d$ is associated with start time $s_k$ and end time $e_k$. 
We define the discourse label space as $\mathcal{Y}^d = \{L_1, \ldots, L_{N^d}\}$, where $N^d$ is the number of discourse labels. 
Each talk turn $x_k^d$ is annotated with zero or more labels from $\mathcal{Y}^d$. 


\subsection{Vision Pathway}\label{sec:vision-pipeline}
To address multi-label activity recognition in real-world classroom videos, we employed and evaluated three general approaches that spanned the spectrum from training-free to supervised methods: (1) zero-shot prompting of MLLMs, (2) fine-tuning of vision-language discriminative models (VLDMs), and (3) attentive probing of self-supervised video transformer encoders.

\paragraph{Video Input with Temporal Context.}
Contextual cues have been demonstrated to be essential to understand human activities~\cite{qing2021temporal,shi2023multiemo,wang2023context}. Inspired by this, we extended the target clip to a $\tau$-second temporal window $w_t^v = \{x_{t-\tau+1}^v, \ldots, x_t^v\}$, enabling models to better capture classroom dynamics. Using preceding seconds as context was aligned with the practical application of providing real-time feedback to teachers through the dashboard. 

\paragraph{Multimodal Large Language Models.}
Through multimodal pretraining objectives and instruction tuning, MLLMs extended LLMs' natural language generation capabilities to visual reasoning, enabling a wide range of downstream vision applications without task-specific training. As such, our work aimed to explore MLLMs' zero-shot ability to identify classroom activities in the wild. We adopted the open-source \textit{Qwen2.5-VL}~\cite{bai2025qwen2} model rather than commercial alternatives like \textit{ChatGPT}, as (1) it enabled local deployment without sharing sensitive data externally, addressing privacy concerns when handling classroom recordings involving minors, and (2) avoided the significant API costs associated with processing video inputs. 

The \textit{Qwen2.5-VL} model took as input the frames from a window $w_t^v$, paired with a Chain-of-Thought (CoT)~\cite{wei2022chain} prompt. The prompt (see Supplementary Material) positioned the model as an expert classroom observer, enumerated each activity label in $\mathcal{Y}^v$ with a short description, and instructed it to describe which activities were present in the scene. The model generated a response in JSON format, representing a list of selected labels, which were then post-processed into a binary vector $\hat{y}_t^v \in \{0,1\}^{N^v}$.

\paragraph{Vision-Language Discriminative Models.}
In addition to MLLMs, another family of VLMs consisted of discriminative models like \textit{CLIP}~\cite{radford2021learning}, which learned a joint embedding space by aligning visual and textual representations through large-scale contrastive pretraining.
For our task, we employed \textit{X-CLIP}~\cite{XCLIP}, which adapted \textit{CLIP}'s language-image alignment framework to video temporal modeling. 

Following \textit{X-CLIP}, a window $w_t^v$ was first encoded into a video embedding by the video encoder $f_{\theta_v}$:
\begin{equation}
    z_v = f_{\theta_v}(w_t^v).
\end{equation}
Similarly, each label name $L_i$ in $\mathcal{Y}^v$ was fed into the text encoder $f_{\theta_t}$:
\begin{equation}
    z_{L_i} = f_{\theta_t}(L_i), i = 1,\ldots,N^v.
\end{equation}
Afterward, the cosine similarity between the video and label $L_i$ was computed as
\begin{equation}
    \text{sim}_i = \frac{z_v z_{L_i}}{\|z_v\|\|z_{L_i}\|}.
\end{equation}
These similarity scores served as logits for multi-label classification. We applied a binary cross-entropy (BCE) loss to maximize similarity scores for present activities at time step $t$ while minimizing scores for absent ones. Class weights were incorporated to address label imbalance.

\paragraph{Video Transformer Encoders.}
In contrast to \textit{X-CLIP}, which formulated activity recognition as a vision-language matching problem, video transformers relied specifically on visual supervision. The effectiveness of pretrained masked autoencoders through self-supervised learning has been demonstrated in downstream video understanding tasks~\cite{tong2022videomae,assran2025v}. 
Building on this line of work, we adopted the state-of-the-art video encoder \textit{V-JEPA 2}~\cite{assran2025v}, capable of producing high-quality video representations by scaling self-supervised pretraining to extensive collections of internet videos. 
For downstream tasks, \textit{V-JEPA 2} trained an attentive probe on top of the frozen encoder to aggregate the generated sequential representation through self- and cross-attention mechanisms. 

We fed a window $w_t^v$ into the frozen encoder $f_{\theta}$, yielding a sequence of spatiotemporal token embeddings: 
\begin{equation}
    H = f_{\theta}(w_t^v) \in \mathbb{R}^{N \times D},
\end{equation}
where $N$ was the number of tokens and $D$ was the embedding dimension. 
Afterward, these embeddings $H$ were passed to a trainable attentive probe, comprised of three self-attention layers and one cross-attention layer over a learnable query token $q$:
\begin{equation}
    H^{(\ell)} = \text{SelfAttn}_\ell\!\left(H^{(\ell-1)}\right), 
    \quad \ell = 1,2,3,
\end{equation}
\begin{equation}
    z = \text{CrossAttn}\!\left(q, H^{(3)}\right) \in \mathbb{R}^D,
\end{equation}
where $H^{(0)} = H$ and $z$ was the pooled representation of the video window $w_t^v$. 
Finally, a linear classification head was stacked on top of the attentive probe to perform multi-label classification, optimized with the weighted BCE loss.




\subsection{Transcript Pathway}\label{sec:transcript-pipeline}
In parallel with the vision pathway, we explored three approaches for multi-label classification of discourse activities in classroom transcripts: (1) zero-shot prompting of LLMs, (2) few-shot prompting of LLMs, and (3) fine-tuning an embedding-based classifier.

\paragraph{Large Language Models.}\label{sec:llms}
As in the vision experiments, we relied on open-source models, since privacy is paramount when working with classroom transcripts. Specifically, we evaluated two variants of instruction-tuned LLMs: \textit{Llama3 Instruct}~\cite{llama3modelcard}, in both its \textit{8B} and \textit{70B} variants.

All models were evaluated with the same prompt structure. For the zero-shot setting, the prompt began with a task description, followed by the complete list of 19 discourse labels and a short explanation. We also included examples of label pairings: some that frequently co-occur (e.g., \textit{Cognitive-Demand\_Analysis-Request} and \textit{Expl-Just\_Teacher-Request}), and others that are mutually exclusive (e.g., \textit{Questions\_Closed-Ended} and \textit{Questions\_Open-Ended}). The speaker information preceded each transcript excerpt, and the models were asked to return a JSON object with all labels and their predictions. The same prompt was augmented with two real transcript examples and their corresponding labels in the few-shot setting. Full prompt templates are provided in the Supplementary Material.

\paragraph{Model Fine-tuning.}\label{modelfinetuning}
Besides prompting LLMs, we fine-tuned a transformer-based model for the multi-label classification task. First, the transcript data were preprocessed to standardize both the speaker annotation and the temporal boundaries of each utterance. To incorporate conversational context, we appended the two preceding utterances, marked with the tokens ${<}ctx_{i-2}{>}$ and ${<}ctx_{i-1}{>}$, to the target utterance, marked by ${<}curr{>}$. Special tokens indicating the beginning and end of each segment were added to the tokenizer. A maximum sequence length of 576 was set, which covered 99\% of all utterances without truncation.

The model architecture consisted of a transformer-based encoder and a single linear classification head. We experimented with several \textit{BERT}-derived models~\cite{devlin-etal-2019-bert}, with \textit{DeBERTaV3}~\cite{he2023debertav3improvingdebertausing} yielding the best performance. \textit{DeBERTaV3} extends \textit{DeBERTa}~\cite{he2021debertadecodingenhancedbertdisentangled} by replacing the Masked Language Modeling objective with Replaced Token Detection and introducing disentangled embedding sharing, an enhanced embedding-sharing mechanism inspired by \textit{ELECTRA}~\cite{clark2020electrapretrainingtextencoders}. The encoder was initialized using the HuggingFace implementation of \textit{DeBERTaV3}, and the \texttt{[CLS]} token representation was used as the sequence embedding. Both the encoder and the classification head were fine-tuned jointly, with no layers frozen.

Given the large number of labels and the severe imbalance, we evaluated three different loss functions: BCE~\cite{Goodfellow-et-al-2016}, Focal Loss~\cite{focalloss}, and Asymmetric Loss (ASL)~\cite{benbaruch2020asymmetric}, with or without positive label weighting.

\paragraph{Binary Cross-Entropy.}  
In the first approach, we used standard BCE but weighted the positive labels to counter class imbalance. For each label $j$, the positive weight is computed as the ratio of negative to positive examples in the training data:
\begin{equation}
    \mathrm{pos\_weight_j} =
    \begin{cases}
        \dfrac{\mathrm{neg\_counts[j]}}{\mathrm{pos\_counts[j]}}, & \text{for positive weighting}, \\[1.2ex]
        1, & \text{otherwise}.
    \end{cases}
\end{equation}
where $\mathrm{neg\_counts[j]}$ and $\mathrm{pos\_counts[j]}$ denote the number of negative and positive instances of label $j$, respectively. The weighted BCE loss for a single label $j$ is then:
\begin{equation}
    \begin{split}
        \ell_j(p,y) = -\Big(&y \cdot \log(\sigma(p) \cdot \mathrm{pos\_weight_j}) \\
                        &+ (1-y) \cdot \log(1-\sigma(p))\Big),
    \end{split}
\end{equation}
where $p$ are the raw logits, $y$ is the gold label, and $\sigma$ denotes the sigmoid function.

\paragraph{Focal Loss.}  
The second approach employed Focal Loss, which down-weights easy-to-classify examples, allowing the model to focus on harder, minority-class instances:
\begin{equation}
    \begin{split}
        \mathrm{FL(p)} = &-\alpha \cdot (p)^\gamma \cdot \log(p) \cdot \mathrm{pos\_weight_j} \\
                &-(1-\alpha) \cdot (1-p)^\gamma \cdot \log(1-p),
    \end{split}
\end{equation}
where $p$ is the predicted probability of the true class ($1-p$ for negatives). The parameter $\alpha \in [0,1]$ balances positive and negative classes, while $\gamma \geq 0$ controls the degree of down-weighting applied to easy examples.

\paragraph{Asymmetric Loss.}  
Finally, we evaluated Asymmetric Loss, which introduces separate focusing parameters for positive and negative classes ($\gamma_+$ and $\gamma_-$). This allows the model to suppress easy negatives ($\gamma_- > 0$) while not suppressing positives ($\gamma_+ \approx 0$). The formulation is:
\begin{equation}
    \begin{split}
        \mathrm{ASL(p, y)} = &-y \cdot (1-p)^{\gamma_+} \cdot \log(p) \cdot \mathrm{pos\_weight_j} \\
                    &-(1-y) \cdot (p_m)^{\gamma_-} \cdot \log(1-p_m),
    \end{split}
\end{equation}
where $p$ is the predicted probability, $y$ is the true label, and $p_m = \max(p - m, 0)$ introduces a margin $m$ for negative examples.

\section{Experiments}
\label{sec:experiments}

\subsection{Dataset Setup and Evaluation Metrics}
\label{sec:datasetsetup}
We adopted a lesson-independent split of the dataset (Sec.~\ref{sec:data}), such that all clips from the same lesson remained within a single partition, enabling the evaluation of models' generalizability to unseen lessons. The resulting split contained 115 hours of videos and 21,208 transcript turns for training, 23 hours and 7,386 turns for validation, and 26 hours and 4,473 turns for testing. 
Given the dataset's class imbalance, we evaluated models using macro-F1 to ensure that both rare and frequent classes were weighted equally.

\subsection{Implementation Details}
\label{sec:implementationdetails}
\paragraph{Vision Pathway.}
For zero-shot prompting MLLMs, we used \textit{Qwen2.5-VL-32B-Instruct} with default generation hyperparameters. 
For finetuning VLDMs, we employed \textit{X-CLIP}~\cite{XCLIP} with both \textit{ViT-B/16}~\cite{dosovitskiy2020image} and \textit{ViT-L/14} backbones, initialized with weights pre-trained on \textit{Kinetics-400}~\cite{carreira2017quo} at a resolution of $224\times224$ and $336\times336$, respectively. Following~\cite{XCLIP}, the video encoder was fine-tuned, while the text encoder was frozen. Training was conducted for 10 epochs using the AdamW optimizer~\cite{loshchilov2017decoupled}, with a batch size of 8 and a learning rate of $8 \times 10^{-6}$.
For the video encoder approach, we explored\textit{ V-JEPA 2}~\cite{assran2025v} with both \textit{ViT-L/16} ($256\times256$ resolution) and \textit{ViT-G/16} ($384\times384$ resolution) backbones. For both backbones, we initialized the attentive probe from weights pretrained from \textit{SSv2}~\cite{goyal2017something} and subsequently finetuned it on our dataset. Optimization followed the same setup for \textit{X-CLIP}, except with a learning rate of $1 \times 10^{-3}$, since only the final attention-wise classifier was trained.
For all approaches, we uniformly sampled each one-second clip $x_t^v$ into $F=8$ frames and, unless otherwise specified, set the temporal window duration to $\tau=2$ seconds.
Additionally, since adjacent clips often exhibit low visual variation, we temporally selected every 5th clip during training and validation to reduce redundancy and computational cost. We used the complete set of clips for testing to ensure a comprehensive evaluation.

\paragraph{Transcript Pathway.}  
For zero-shot and few-shot prompting of LLMs, we evaluated \textit{Llama3 Instruct}~\cite{llama3modelcard} (8B, 70B). 
For fine-tuning, we employed \textit{DeBERTaV3}~\cite{he2023debertav3improvingdebertausing}, initialized from HuggingFace pretrained weights, with a single linear classification head. Both encoder and classifier were fine-tuned jointly. 
Training was performed for up to 30 epochs, with early stopping after 10 consecutive epochs with no improvement, using AdamW~\cite{loshchilov2017decoupled} with learning rate $1\times 10^{-5}$, batch size 32, warmup ratio 0.1, weight decay 0.01, dropout 0.1, and a cosine scheduler. 

\paragraph{Decision Thresholding.}  
During evaluation, both the vision and transcript models output probabilities for each label using the sigmoid of the logits. Since these probabilities must be binarized, we searched for optimal per-label thresholds instead of applying a fixed cutoff (e.g., 0.5). Specifically, for each label $j$, candidate thresholds $T_j \in [0, 1]$ were tested in increments of 0.05, and the value that maximized macro-F1 on the validation set was selected. The best-performing set of thresholds from the best validation epoch was then applied to the test set.


\subsection{Main Results}
\label{sec:results}

\paragraph{Vision Pathway.} Table~\ref{tab:video_results} presents the performance of the investigated approaches for instructional activity recognition in classroom videos. 
The zero-shot prompting of \textit{Qwen2.5-VL} achieved a macro-F1 of 0.423, indicating the potential of MLLMs to transfer some generalizable knowledge to classroom activity understanding. 
Nevertheless, its performance and throughput consistently lagged behind fine-tuned approaches, reflecting the value of domain-specific adaptation.
Among task-specialized methods, \textit{X-CLIP} with \textit{ViT-L/14} backbone resulted in the best overall performance, with a macro-F1 of 0.577. Notably, it outperformed the larger \textit{V-JEPA 2 ViT-G/16} by 0.03, suggesting natural language supervision through label descriptions provided beneficial semantic guidance for our real-world activity recognition task. It also outperformed the BaS-Net model from Foster et al.~\cite{foster2024automated}, using the same dataset and labels.
Moreover, the consistent improvements from larger model variants within each family, i.e., \textit{X-CLIP ViT-L/14} over \textit{ViT-B/16}, and \textit{V-JEPA 2 ViT-G/16} over \textit{ViT-L/16}, indicated that increased model capacity contributed to a better understanding of classroom dynamics.

\begin{table*}
    \centering
    \footnotesize
    \begin{tabular}{lllllll}
        \toprule
        \textbf{Method} & \textbf{Model} & \textbf{Variant} & \textbf{\#Params} & \textbf{Resolution} & \textbf{Throughput} & \textbf{Macro-F1} \\
        \midrule
        Zero-shot & Qwen2.5-VL & 32B-Inst. & 32B & dynamic & 0.07 & 0.423 \\
        \midrule
        \multirow{4}{*}{Fine-tuning}
         & \multirow{2}{*}{X-CLIP} & ViT-B/16 & 0.2B & 224 & 29.21 & 0.513 \\
         &                         & ViT-L/14 & 0.6B & 336 & 9.09 & 0.577 \\
        \cmidrule{2-7}
        & \multirow{2}{*}{V-JEPA 2} & ViT-L/16 & 0.4B & 256 & 14.21 & 0.525 \\
        &                         & ViT-G/16 & 1.1B & 384 & 6.55 & 0.547 \\
        \midrule
        Foster et al.~\cite{foster2024automated} & BaS-Net & -- & -- & --
        & -- & 0.469\\
        \bottomrule
    \end{tabular}
    \caption{Performance of different approaches for instructional activity recognition in classroom videos. All methods were evaluated with a temporal context window of 2 seconds (i.e., 16 frames in total) as input. Throughput (clips/s) was measured on an A100 GPU. Foster et al.~\cite{foster2024automated} was added as comparison, since they worked with the same dataset and labels.}
    \label{tab:video_results}
\end{table*}

\paragraph{Transcript Pathway.} Table~\ref{tab:disc_res_loss} presents the performance of the investigated approaches for instructional discourse recognition in classroom transcripts.
Few-shot prompting with \textit{Llama3} improved slightly over the zero-shot setting, but both variants achieved very low macro-F1 ($\leq0.079$), highlighting the difficulty of applying LLM prompting to this multi-label task and their tendency to generate inconsistent outputs.
In contrast, fine-tuned \textit{DeBERTaV3} models achieved substantially higher scores, underscoring the effectiveness of task-specific adaptation.
Among the evaluated loss functions, Asymmetric Loss with positive-label weighting (PW) yielded the best performance with a macro-F1 of 0.460, confirming its suitability for rare-label prediction in imbalanced multi-label classification.
Notably, while positive weighting improved results for both BCE and ASL, it degraded performance for Focal Loss. This was likely due to the interaction between its focusing term and additional weighting, which may have overemphasized positives, harmed calibration, and reduced precision.

\begin{table}[ht]
    \centering
    \footnotesize
    \begin{tabular}{llll}
    \toprule
    \textbf{Method} & \textbf{Model} & \textbf{Variant} & \textbf{Macro-F1} \\
    \midrule
    \multirow{2}{*}{Zero-shot} 
     & \multirow{2}{*}{LLaMA}   & 8B Inst.      & 0.060 \\
     &                          & 70B Inst.     & 0.091 \\ 
    \midrule
    \multirow{2}{*}{Few-shot} 
     & \multirow{2}{*}{LLaMA}   & 8B Inst.      & 0.079 \\
     &                          & 70B Inst.     & 0.265 \\ 
    \midrule
    \multirow{6}{*}{Fine-tuning} 
     & \multirow{6}{*}{DeBERTaV3} 
       & BCE        & 0.422 \\
     & & BCE + PW   & 0.441 \\ \cline{3-4}
     & & Focal      & 0.434 \\
     & & Focal + PW & 0.428 \\ \cline{3-4}
     & & ASL        & 0.403 \\
     & & ASL + PW   & \textbf{0.460} \\
    \bottomrule
    \end{tabular}
    \caption{Macro-F1 scores for different models and loss functions. BCE, Focal, and ASL refer to Binary Cross-Entropy loss, Focal loss, and Asymmetric loss, respectively. These losses were used with or without weighting of the positive labels (PW).}
    \label{tab:disc_res_loss}
\end{table}


\paragraph{Effect of temporal context.} To examine the role of temporal context in multi-label classroom activity recognition, we compared performance across different contextual lengths using the best-performing \textit{X-CLIP ViT-L/14} model. As shown in Table~\ref{tab:temporal_context}, adding one second of prior context improved macro-F1 from 0.558 to 0.577 over using the target clip alone, while extending the window to three seconds yielded a slight drop of 0.01. For discourse recognition, we tested the best-performing \textit{DeBERTaV3} model with and without two preceding utterances. In this case, removing context caused a substantial decline in macro-F1 (0.460 to 0.392). These findings suggested that short-term temporal cues are beneficial for visual activity recognition, while sentence-level context is crucial for discourse modeling. However, excessively long windows may introduce irrelevant dynamics that hinder fine-grained prediction.

\begin{table}
    \centering
    \footnotesize
    \begin{tabular}{llll}
        \toprule
        &\textbf{Input Setting} & \textbf{\#Frames} & \textbf{Macro-F1} \\
        \midrule
        \multirow{3}{2em}{\textbf{Video}}&Target clip only & 8 & 0.558 \\
        &\hspace{0.4cm} +1s context & 16 & 0.577 \\
        &\hspace{0.4cm} +3s context & 32 & 0.567 \\
        \midrule
        \multirow{2}{2em}{\textbf{Text}} & Target sentence only  & -  & 0.392 \\
        & \hspace{0.4cm} +2 sent. context & -  & 0.460 \\
        \midrule
    \end{tabular}
    \caption{Results of video-based instructional activity recognition (X-CLIP ViT-L/14) and text-based instructional discourse recognition (DeBERTaV3 Embedding Classifier with Asymmetric loss and positive label weighting) across different temporal and textual context lengths.}
    \label{tab:temporal_context}
\end{table}



\section{Discussion}
\label{sec:discussion}
Our experiments demonstrate the feasibility of applying advanced vision and language models to the recognition of instructional practices in authentic classroom data. Across both modalities, domain-specific fine-tuning consistently outperforms prompting-based approaches, while incorporating short-term temporal context improves recognition of classroom activities. These findings highlight the need to tailor model architectures and training strategies to the multimodal, imbalanced, and noisy nature of classroom data in order to effectively transfer advances in vision and language modeling to the educational domain.

The demonstrated feasibility of automated classroom analysis opens several avenues for educational applications. The macro-F1 scores achieved (0.577 for video, 0.460 for transcripts) represent meaningful progress but fall short of the reliability required for real-world deployment. Instead, these systems are better positioned as supplementary tools for teacher professional growth and large-scale research, with pilot studies involving educators playing a key role in validating their pedagogical usefulness. For instance, automatic recognition of small-group discussions or extended teacher explanations could be aggregated into time-based summaries, allowing teachers to reflect on the balance between student- and teacher-led activities. Similarly, identifying frequent transitions between activities could highlight classroom management patterns. Such outputs provide concrete and interpretable signals that can be translated into actionable feedback for instructional practice.

Several limitations warrant consideration. First, our approach relies on parallel, modality-specific pipelines rather than joint fusion models, due to the nature of the annotations, where activity labels are derived only from videos and discourse from transcripts. Initial attempts at fusion—either training a single model to predict all 43 labels or applying weighted fusion between video and transcript classifiers—underperformed compared to unimodal pipelines. Further exploration of effective fusion mechanisms remains an important avenue for future research. Second, recognition performance is uneven across labels: low-frequency practices such as \textit{on task student talking with student} (visual) and \textit{student request} (discourse) remain particularly challenging. This reflects both class imbalance and the complexity of subtle, interaction-rich practices, underscoring the need for methods that better handle rare classes, such as data augmentation or few-shot learning. Third, practical deployment raises throughput constraints. The best-performing \textit{X-CLIP ViT-L/14} processes roughly 9 clips/s on high-end GPUs, which may be insufficient for real-time classroom feedback. Bridging this gap between research-grade models and lightweight, cost-effective solutions for schools will require approaches such as knowledge distillation or streaming architectures. Finally, our models are trained solely on elementary mathematics and ELA classrooms. Therefore, generalization to other subjects, grade levels, and diverse student–teacher demographics requires further validation.

Applying AI to classroom recordings raises serious ethical concerns, particularly when processing sensitive data involving entire classrooms of minors~\cite{buhler2023automated,sumer2021multimodal,hou2025multimodalassessmentclassroomdiscourse}. Our choice to rely on open-source models deployable locally mitigates some privacy risks, but other concerns remain. Models may exhibit algorithmic biases, performing differently across teaching styles, student demographics, or cultural contexts. Moreover, transparency and informed consent are critical: students and teachers must be aware of how their data is used and retain the option to opt out. Ultimately, AI-driven systems should provide interpretable, actionable guidance to teachers, who remain responsible for final judgments and feedback, ensuring that technology supports rather than replaces human expertise.
\section{Conclusion}
\label{sec:conclusion}
This work demonstrates that advanced vision and language models can be adapted to recognize multimodal instructional practices from authentic classroom data, with fine-tuned pipelines clearly outperforming zero- and few-shot prompting approaches. By systematically benchmarking video- and transcript-based methods on a large, lesson-independent dataset, we provide the first evidence that domain-specific adaptation is crucial for robust instructional recognition. While current performance does not yet support high-stakes evaluation, our results establish a foundation for scalable, automated teacher feedback systems that can augment professional development and large-scale educational research. Moving forward, advances in multimodal fusion, generalization to broader contexts, and ethically grounded deployment will be key to transforming these technical capabilities into trustworthy and pedagogically meaningful classroom tools.

\paragraph{Acknowledgments.}
This project was funded by the Hector Foundation as part of the Hector AI+ Education Future Fund.

{
    \small
    \bibliographystyle{ieeenat_fullname}
    \bibliography{main}
}

\clearpage
\appendix

\definecolor{frame}{HTML}{ccefff}    
\definecolor{back}{HTML}{FAFEFF}     

\definecolor{t_frame}{HTML}{ccefff}    
\definecolor{t_back}{HTML}{ccefff}     

\section{Instructional Labels}
\label{app:labels}
The dataset contains 43 labels: 24 instructional activity labels (see Table~\ref{tab:actlabels}) and 19 instructional discourse labels (see Table~\ref{tab:dislabels}). For the discourse labels, the dataset includes information of labels that are usually paired together: 
\begin{itemize}
    \item \textbf{Analysis Request (Teacher)} and \textbf{Explanation/Justification Teacher Request}.
    \item \textbf{Analysis Give (Teacher)} and \textbf{Explanation/Justification Teacher Give}.
    \item \textbf{Feedback Neutral}, \textbf{Uptake Restating}, and \textbf{Feedback Unelaborated}.
\end{itemize}

\rowcolors{2}{gray!15}{white} 

\begin{table*}[t]
    \centering
    \begin{tabularx}{\textwidth}{|p{6cm}|X|}
    \hline
    \textbf{Instructional Activity Label} & \textbf{Description} \\
    \hline
     Whole class activity & All students are involved in one activity, with the teacher leading the learning (e.g., lecture, presentation, carpet time). \\
     Individual activity & Students work independently (e.g., practice, reading), no interaction with peers. \\
     Small group activity & Students working together with peers (e.g., think-pair-share, book club). \\
     Transition & Teacher/students move between activities or locations; no meaningful instruction. \\
     On-task student talking with student & Students conversing with each other without teacher support; can overlap with small group activity. \\
     Student raising hand & Hand up for more than 1 second, clearly and purposefully. \\
     Teacher sitting & Teacher seated (chair, stool, floor, crouching, kneeling, on desk). \\
     Teacher standing & Teacher standing in the same spot/orientation to students. \\
     Teacher walking & Teacher walking with purpose to change orientation. \\
     Students sitting on carpet or floor & Students seated on the carpet/floor. \\
     Students sitting at group tables & Students seated at tables. \\
     Sitting at desks & Students seated at individual desks. \\
     Students standing or walking & One or more students standing or moving around. \\
     Teacher supporting one student & Teacher assists one student verbally or non-verbally. \\
     Teacher supporting multiple students with interaction & Teacher assists multiple students who are also interacting with one another. \\
     Teacher supporting multiple students without interaction & Teacher assists multiple students who are not interacting with one another. \\
     Using or holding book & Book is used/held by teacher or student. \\
     Using or holding worksheet & Worksheet is used/held by teacher or student. \\
     Presentation with technology & Smartboard, Elmo, projector used to show content. \\
     Using or holding instructional tool & Object (e.g., ruler, manipulative) used/held for instruction (not pen/pencil/furniture). \\
     Using or holding notebook & Notebook is used/held by teacher or student. \\
     Individual technology & Laptop, tablet, or other personal tech used by student or teacher. \\
     Teacher writing & Teacher writes/erases on paper, board, or document camera. \\
     Student writing & Student writes/erases on paper or board. \\
    \hline
    \end{tabularx}
    \caption{Instructional activity labels.}
    \label{tab:actlabels}
\end{table*}

\begin{table*}[t]
    \centering
    \begin{tabularx}{\textwidth}{|p{6cm}|X|}
    \hline
    \textbf{Instructional Discourse Labels} & \textbf{Description} \\
    \hline
     Analysis Give (Teacher) & Teacher analyzes content (e.g., compare, justify, synthesize, connect). \\
     Analysis Request (Teacher) & Teacher asks students to analyze content (e.g., why/how, connect ideas). \\
     Report Give (Teacher) & Teacher states facts, definitions, or procedures. \\
     Report Request (Teacher) & Teacher asks students to recall or report facts, definitions, or methods. \\
     Questions Open-Ended & Teacher asks open content-related questions without a pre-scripted answer. \\
     Questions Closed-Ended & Teacher asks content-related questions with a pre-scripted/fluency answer (e.g., yes/no). \\
     Questions Task Related Prompt & Teacher reads/restates a task question/prompt from instructional materials. \\
     Explanation/Justification Teacher Request & Teacher requests an explanation or justification. \\
     Explanation/Justification Teacher Give & Teacher provides an explanation or justification, possibly a solution strategy. \\
     Explanation/Justification Student Request & Student requests an explanation or justification. \\
     Explanation/Justification Student Give & Student provides an explanation or justification. \\
     Feedback Affirming & Teacher positively evaluates student’s response (e.g., “Excellent”). \\
     Feedback Disconfirming & Teacher negatively evaluates correctness of a response (e.g., “No”). \\
     Feedback Neutral & Teacher neither confirms nor disconfirms, may repeat student’s contribution. \\
     Feedback Elaborated & Teacher expands on student’s response or thinking. \\
     Feedback Unelaborated & Teacher acknowledges, evaluates, or superficially uses a student’s idea. \\
     Uptake Restating & Teacher repeats/summarizes a student’s idea without building on it. \\
     Uptake Building & Teacher incorporates/clarifies or expands on a student’s idea. \\
     Uptake Exploring & Teacher probes further into a student’s idea with follow-up questions. \\
    \hline
    \end{tabularx}
    \caption{Instructional discourse labels.}
    \label{tab:dislabels}
\end{table*}

\section{Label Distribution}
\label{app:labeldist}

Table~\ref{tab:positivelabeldist} reports the distribution of positive labels for both Instructional Activity and Instructional Discourse categories. For Activity labels, annotations are provided at the per-second level of the video; thus, the percentages reflect the proportion of seconds in which a given label is active. In contrast, Discourse labels are annotated at the utterance level. To enable comparison across modalities, we report both the proportion of utterances containing each label (\% Utt) and a time-based measure showing the proportion of seconds in which the label occurs (\% Sec).

The distributions highlight the strong imbalance across labels. For instance, highly frequent activity labels such as Teacher standing (52.4\%) and Whole class activity (51.9\%) dominate the dataset, while more fine-grained instructional practices like Teacher writing (4.1\%) or On-task student talking with student (4.2\%) are comparatively rare. A similar skew is evident in the discourse dimension: labels such as Report-Request appear in 20.5\% of utterances, whereas others like Analysis-Give (0.5\%) and Student-Request (0.2\%) occur only sporadically. This imbalance underscores the difficulty of reliably detecting rare but pedagogically important practices, and motivates our exploration of imbalance-aware objectives and dynamic thresholding strategies.

\begin{table*}
    \centering
    \begin{tabular}{p{7cm}p{1cm}  p{5cm}p{1cm}p{1cm}}
    \toprule
    \textbf{Activity Labels} & \textbf{\% Sec} & \textbf{Discourse Labels} & \textbf{\% Utt} & \textbf{\% Sec} \\
    \midrule
    Teacher standing & 52.4 & Report-Request & 20.5 & 13.4 \\
    Whole class activity & 51.9 & Closed-Ended & 13.4 & 7.8 \\
    Presentation with technology & 44.5 & Feedback-Neutral & 10.0 & 6.0 \\
    Sitting at desks & 43.2 & Feedback-Unelaborated & 10.8 & 6.0 \\
    Using or holding worksheet & 40.6 & Open-Ended & 9.1 & 6.0 \\
    Students sitting at group tables & 33.1 & Feedback-Elaborated & 4.9 & 4.0 \\
    Students standing or walking & 30.3 & Uptake-Restating & 5.6 & 3.5 \\
    Students sitting on carpet or floor & 29.0 & Feedback-Affirming & 4.9 & 3.3 \\
    Teacher sitting & 26.8 & Uptake-Building & 3.1 & 2.6 \\
    Small group activity & 22.1 & Report-Give & 2.3 & 2.3 \\
    Student writing & 19.2 & Analysis-Request & 3.3 & 2.1 \\
    Using or holding notebook & 18.7 & Teacher-Request & 2.6 & 1.4 \\
    Individual activity & 17.6 & Uptake-Exploring & 3.9 & 1.4 \\
    Using or holding instructional tool & 17.3 & Task Related Prompt & 0.7 & 1.4 \\
    Individual technology & 14.0 & Teacher-Give & 0.9 & 1.3 \\
    Teacher supporting multiple students with student interaction & 13.9 & Analysis-Give & 0.5 & 0.8 \\
    Using or holding book & 13.3 & Feedback-Disconfirming & 0.8 & 0.7 \\
    Teacher walking & 8.9 & Student-Give & 2.3 & 0.7 \\
    Student raising hand & 6.5 & Student-Request & 0.2 & 0.1 \\
    Teacher supporting one student & 6.5 & & & \\
    Transition & 5.7 & & & \\
    Teacher supporting multiple students without student interaction & 4.8 & & & \\
    On task student talking with student & 4.2 & & & \\
    Teacher writing & 4.1 & & & \\
    \bottomrule
    \end{tabular}
    \caption{Percentages of positive and negative labels for Activity and Discourse. For Discourse labels, we report both the proportion of utterances containing each label (\% Utt) and the proportion of seconds in which the label occurs (\% Sec), consistent with the time-based reporting used for Activity labels.}
    \label{tab:positivelabeldist}
\end{table*}

\section{Language Models Prompts}
\label{app:prompts}

\subsection{Multimodal Large Language Models}
\label{app:mllmprompts}
Figure~\ref{fig:mllmzero} shows the prompt for zero-shot for MLLMs.

\begin{figure*}
    \centering
    \begin{tcolorbox}[colback=back, colframe=frame, boxrule=3pt, left=0.5em, right=0.5em, top=0.5em, bottom=0.5em,]
    \# Task Description
    
    You are an expert classroom activity observer. You are given a one-second video clip from a classroom lesson. Carefully analyze the clip, think step-by-step, and identify all instructional activity labels that apply. You may select multiple labels. Each label is listed below with its definition to guide your judgment.
    
    \vspace{3pt}
    
    \# Instructional Activity Labels and Definitions
    
    \hspace{10pt} \{\textit{Names and definitions of 24 instructional activity labels}\}
    
    \vspace{3pt}
    
    \# Instructions
    
    Always return your answer as a JSON list containing the label names that exactly match those provided above (verbatim). If no labels apply, return an empty list. Do not include descriptions, explanations, commentary, or any modified text.
    \end{tcolorbox}

    \caption{Prompt for multimodal large language models. \{\textit{Names and definitions of 24 instructional activity labels}\} can be found in Section~\ref{app:labels}.}
    \label{fig:mllmzero}
\end{figure*}

\subsection{Large Language Models}
\label{app:llmprompts}
Figure~\ref{fig:llmzero} shows the prompt for zero-shot and Figure~\ref{fig:llmfew} show the prompt for few-shot for LLMs.

\begin{figure*}
    \centering
    \begin{tcolorbox}[colback=back, colframe=frame, boxrule=3pt, left=0.5em, right=0.5em, top=0.5em, bottom=0.5em,]
    \#\#\# Task Description: 
    For each of the following 19 categories, output 1 if that teaching activity appears in the given transcript excerpt, otherwise 0. Take into account who is the speaker and context of the transcript excerpt. Output exactly one JSON object with all 19 keys.
    
    \vspace{10pt}
    
    \#\#\# Categories:
    
   \hspace{10pt} \{\textit{Names and definitions of 19 instructional discourse labels}\}
    
    \vspace{10pt}
    
    \#\#\# Further Instructions:
    
    If you're unsure, assign 0 rather than inventing a 1.
    
    Cognitive-Demand\_Analysis-Request and Expl-Just\_Teacher-Request are usually paired together.
    
    Cognitive-Demand\_Analysis-Give and Expl-Just\_Teacher-Give are usually paired together.
    
    Feedback\_Neutral, Feedback\_Unelaborated and Uptake\_Restating are usually paired together.
    
    Feedback\_Affirming, Feedback\_Disconfirming and Feedback\_Neutral are mutually exclusive.
    
    Feedback\_Elaborated and Feedback\_Unelaborated are mutually exclusive.
    
    Uptake\_Building, Uptake\_Exploring and Uptake\_Restating are mutually exclusive.
    
    Questions\_Closed-Ended and Questions\_Open-Ended are mutually exclusive.
    
    Output exactly one JSON object, no extra keys, no explanations.
    
    Ensure valid JSON: double-quoted keys, commas between pairs, no trailing comma.
    
    \vspace{10pt}
    
    \#\#\# Task:
    
    \#\# Context:
    
    \hspace{10pt} \{\textit{Context sentence t-2}\}
    
    \hspace{10pt} \{\textit{Context sentence t-1}\}
    
    \vspace{3pt}
    
    \#\# Speaker:
    
    \hspace{10pt} \{\textit{Current speaker}\}
    
    \vspace{3pt}
    
    \#\# Transcript excerpt:
    
    \hspace{10pt} \{\textit{Current sentence}\}
    
    \vspace{3pt}
    
    \#\# Result:
    \end{tcolorbox}

    \caption{Prompts used for zero-shot prompting of \textit{Llama3.1-Instruct} (8B and 70B).
    \{\textit{Names and definitions of 19 instructional discourse labels}\} can be found in Section~\ref{app:labels}, \{\textit{Context sentence t-2}\} and \{\textit{Context sentence t-1}\} are the two previous sentences, \{\textit{Current speaker}\} is the speaker of the current sentence, and \{\textit{Current sentence}\} is the sentence to be classified.}
    \label{fig:llmzero}
\end{figure*}

\begin{figure*}
    \centering
    \begin{tcolorbox}[colback=back, colframe=frame, boxrule=3pt, left=0.5em, right=0.5em, top=0.5em, bottom=0.5em,]
    \fontsize{10}{9}\selectfont
    \#\#\# Task Description: 
    For each of the following 19 categories, output 1 if that teaching activity appears in the given transcript excerpt, otherwise 0. Take into account who is the speaker and context of the transcript excerpt. Output exactly one JSON object with all 19 keys.
    
    \vspace{10pt}
    
    \#\#\# Categories:
    
   \hspace{10pt} \{\textit{Names and definitions of 19 instructional discourse labels}\}
    
    \vspace{10pt}
    
    \#\#\# Further Instructions:
    
    If you're unsure, assign 0 rather than inventing a 1.
    
    Cognitive-Demand\_Analysis-Request and Expl-Just\_Teacher-Request are usually paired together.
    
    Cognitive-Demand\_Analysis-Give and Expl-Just\_Teacher-Give are usually paired together.
    
    Feedback\_Neutral, Feedback\_Unelaborated and Uptake\_Restating are usually paired together.
    
    Feedback\_Affirming, Feedback\_Disconfirming and Feedback\_Neutral are mutually exclusive.
    
    Feedback\_Elaborated and Feedback\_Unelaborated are mutually exclusive.
    
    Uptake\_Building, Uptake\_Exploring and Uptake\_Restating are mutually exclusive.
    
    Questions\_Closed-Ended and Questions\_Open-Ended are mutually exclusive.
    
    Output exactly one JSON object, no extra keys, no explanations.
    
    Ensure valid JSON: double-quoted keys, commas between pairs, no trailing comma.
    
    \vspace{10pt}

    \#\#\# Example 1:

    \#\# Context:
    
    Teacher 1 (14:35): What about the boy? If you had to describe the boy, how would you describe him? [Student 8]'s hand shot up. [Student 8]. How would you describe him?
    
    Student 8 (14:41): Greedy.
    
    \vspace{3pt}
    
    \#\# Speaker:
    
    Teacher 1

    \vspace{3pt}
    
    \#\# Transcript excerpt:
    
    Teacher 1 (14:42): Greedy? Why do you say greedy?

    \vspace{3pt}

    \#\# Result:

    \hspace{10pt} \{\textit{Valid JSON}\}

    \vspace{10pt}
    
    \#\#\# Example 2:

    \#\# Context:
    
    Student 14 (16:07): And I know that four plus five is nine.
    
    Teacher (16:08): Okay, so you knew that was nine. Student 5 turn around.
    
    \vspace{3pt}
    
    \#\# Speaker:
    
    Student 14

    \vspace{3pt}
    
    \#\# Transcript excerpt:
    
    Student 14 (16:16): I know that 10 plus seven is 17, so I put five and four together and it made nine. And then six and then [00:16:30] that would make 16.
    
    \#\# Result:

    \hspace{10pt} \{\textit{Valid JSON}\}

    \vspace{10pt}
    
    \#\#\# Task:
    
    \#\# Context:
    
    \hspace{10pt} \{\textit{Context sentence t-2}\}
    
    \hspace{10pt} \{\textit{Context sentence t-1}\}
    
    \vspace{3pt}
    
    \#\# Speaker:
    
    \hspace{10pt} \{\textit{Current speaker}\}
    
    \vspace{3pt}
    
    \#\# Transcript excerpt:
    
    \hspace{10pt} \{\textit{Current sentence}\}
    
    \vspace{3pt}
    
    \#\# Result:
    \end{tcolorbox}

    \caption{Prompts used for few-shot prompting of \textit{Llama3.1-Instruct} (8B and 70B). 
    \{\textit{Names and definitions of 19 instructional discourse labels}\} can be found in Section~\ref{app:labels}, \{\textit{Valid JSON}\} is a valid JSON output for the example, here shortened for space, \{\textit{Context sentence t-2}\} and \{\textit{Context sentence t-1}\} are the two previous sentences, \{\textit{Current speaker}\} is the speaker of the current sentence, and \{\textit{Current sentence}\} is the sentence to be classified.}
    \label{fig:llmfew}
\end{figure*}

\section{F1 Score Per Label}
\label{app:f1perlabel}

In this section, we report per-label F1 scores. Table~\ref{tab:f1perlabelact} presents results for Instructional Activity labels using MLLMs and fine-tuned models, while Table~\ref{tab:f1perlabeldisc} shows results for Instructional Discourse labels using LLMs and fine-tuned models. Activity scores were computed on a per-second basis, and discourse scores on a per-utterance basis.
Some labels are clearly more difficult to predict than others. For instructional activities, the best-performing model struggled with labels such as On-task student talking with student (0.080 macro-F1) and Teacher supporting multiple students without student interaction (0.169), while labels like Teacher standing (0.877) and Presentation with technology (0.870) achieved much higher scores. A similar pattern is seen in discourse recognition: Analysis-Give (0.115) and Teacher-Give (0.235) were among the lowest, while Feedback-Affirming (0.625) and Uptake-Restating (0.556) ranked highest. These discrepancies reflect both the rarity of certain labels (see Section~\ref{app:labeldist}) and the inherent difficulty of modeling complex, fine-grained instructional practices.

\begin{table*}[ht]
    \centering
\begin{tabular}{p{4.5cm}ccccccc}
\hline
\textbf{Label} & \textbf{Qwen2.5-VL} & \multicolumn{2}{c}{\textbf{X-CLIP}} & \multicolumn{2}{c}{\textbf{V-JEPA 2}} & \textbf{Foster et al. (2024)} \\
\cline{2-7}
 & 32B-Instruct & ViT-B/16 & ViT-L/14 & ViT-L/16 & ViT-G/16 & \\
\hline
On task student talking with student     & \textbf{0.163} & 0.088 & 0.080 & 0.114 & 0.080 & 0.12  \\
Student raising hand                     & 0.640 & 0.549 & 0.668 & 0.670 & \textbf{0.705} & 0.36  \\
Individual technology                    & 0.698 & 0.520 & \textbf{0.774} & 0.553 & 0.612 & 0.35  \\
Presentation with technology             & 0.851 & 0.865 & \textbf{0.870} & 0.798 & 0.840 & 0.63  \\
Student writing                          & 0.144 & 0.501 & \textbf{0.592} & 0.502 & 0.563 & 0.43  \\
Teacher writing                          & 0.423 & 0.284 & \textbf{0.501} & 0.431 & 0.454 & 0.23  \\
Using or holding book                    & \textbf{0.663} & 0.234 & 0.521 & 0.347 & 0.502 & 0.46  \\
Using or holding instructional tool      & 0.098 & 0.406 & 0.437 & 0.314 & \textbf{0.443} & 0.40  \\
Using or holding notebook                & 0.384 & 0.374 & \textbf{0.390} & 0.315 & 0.315 & 0.24  \\
Using or holding worksheet               & 0.388 & 0.599 & \textbf{0.603} & 0.602 & 0.646 & 0.56  \\
Sitting at desks                         & 0.363 & 0.713 & \textbf{0.763} & 0.722 & 0.762 & 0.74  \\
Students sitting at group tables         & 0.572 & 0.617 & 0.575 & 0.590 & 0.524 & \textbf{0.72}  \\
Students sitting on carpet or floor      & 0.429 & 0.776 & \textbf{0.845} & 0.798 & 0.760 & 0.64  \\
Students standing or walking             & 0.373 & 0.719 & 0.762 & 0.776 & \textbf{0.788} & 0.62  \\
Teacher sitting                          & 0.638 & 0.747 & 0.759 & 0.592 & 0.755 & \textbf{0.78}  \\
Teacher standing                         & 0.820 & 0.844 & 0.877 & 0.875 & \textbf{0.894} & 0.64  \\
Teacher walking                          & 0.014 & 0.465 & 0.538 & 0.606 & \textbf{0.624} & 0.34  \\
Teacher supporting multiple students with student interaction & 0.472 & 0.522 & 0.426 & 0.461 & 0.446 & \textbf{0.54}  \\
Teacher supporting multiple students without student interaction & 0.202 & 0.157 & 0.169 & 0.175 & 0.091 & \textbf{0.40}  \\
Teacher supporting one student           & 0.067 & 0.197 & \textbf{0.286} & 0.190 & 0.269 & 0.27  \\
Individual activity                      & 0.165 & 0.299 & \textbf{0.491} & 0.299 & 0.299 & 0.38  \\
Small group activity                     & 0.627 & \textbf{0.652} & 0.633 & 0.610 & 0.528 & 0.55  \\
Transition                               & 0.106 & 0.371 & \textbf{0.426} & 0.423 & 0.386 & 0.38  \\
Whole class activity                     & 0.854 & 0.817 & \textbf{0.864} & 0.834 & 0.832 & 0.37  \\
\hline
\textbf{Macro F1}                        & 0.423 & 0.513 & \textbf{0.577} & 0.525 & 0.547 & 0.469 \\
\hline
\end{tabular}

    \caption{Per-label F1 scores for instructional activity recognition in videos using different models.}
    \label{tab:f1perlabelact}
\end{table*}

\begin{table*}[ht]
    \centering
    \begin{tabular}{lcccc|cccccc}
        \hline
        \textbf{Label} 
        & \multicolumn{2}{c}{\textbf{Llama 8B}} 
        & \multicolumn{2}{c|}{\textbf{Llama 70B}} 
        & \multicolumn{2}{c}{\textbf{BCE}} 
        & \multicolumn{2}{c}{\textbf{Focal}} 
        & \multicolumn{2}{c}{\textbf{Asym}} \\
        \cline{2-11}
         & Zero & Few & Zero & Few & - & PW & - & PW & - & PW \\
        \hline
        Feedback-Affirming     & 0.093 & 0.192 & 0.096 & 0.405 & 0.604 & 0.571 & 0.594 & 0.576 & 0.516 & \textbf{0.625} \\
        Feedback-Neutral       & 0.161 & 0.161 & 0.095 & 0.348 & \textbf{0.659} & 0.634 & 0.628 & \textbf{0.659} & 0.612 & 0.654 \\
        Feedback-Disconfirming & 0.042 & 0.100 & 0.018 & 0.234 & 0.294 & 0.220 & 0.277 & 0.299 & 0.084 & \textbf{0.322} \\
        Feedback-Elaborated    & 0.089 & 0.101 & 0.049 & 0.275 & \textbf{0.463} & 0.420 & 0.444 & 0.451 & 0.387 & 0.459 \\
        Feedback-Unelaborated  & 0.190 & 0.142 & 0.088 & 0.344 & 0.581 & 0.603 & 0.588 & \textbf{0.618} & 0.586 & 0.586 \\
        Uptake-Building        & 0.000 & 0.000 & 0.061 & 0.213 & 0.383 & 0.356 & 0.384 & 0.369 & 0.338 & \textbf{0.387} \\
        Uptake-Exploring       & 0.052 & 0.124 & 0.065 & 0.394 & 0.406 & 0.375 & 0.370 & 0.364 & 0.296 & \textbf{0.453} \\
        Uptake-Restating       & 0.100 & 0.173 & 0.075 & 0.329 & 0.543 & 0.554 & 0.514 & 0.534 & 0.511 & \textbf{0.556} \\
        Report-Give            & 0.000 & 0.037 & 0.064 & 0.163 & 0.242 & 0.348 & 0.296 & \textbf{0.376} & 0.333 & 0.340 \\
        Report-Request         & 0.000 & 0.000 & 0.031 & 0.176 & \textbf{0.765} & 0.763 & 0.742 & 0.763 & 0.760 & 0.740 \\
        Analysis-Give          & 0.000 & 0.093 & 0.000 & 0.171 & 0.065 & 0.140 & 0.176 & 0.143 & \textbf{0.214} & 0.115 \\
        Analysis-Request       & 0.049 & 0.086 & 0.084 & 0.256 & 0.511 & 0.511 & 0.439 & 0.480 & \textbf{0.517} & 0.480 \\
        Teacher-Give           & 0.007 & 0.000 & 0.022 & 0.223 & 0.071 & 0.214 & 0.225 & 0.154 & \textbf{0.294} & 0.235 \\
        Teacher-Request        & 0.050 & 0.108 & 0.071 & 0.244 & 0.502 & 0.500 & 0.464 & 0.493 & 0.482 & \textbf{0.505} \\
        Student-Give           & 0.064 & 0.074 & 0.063 & 0.337 & \textbf{0.571} & 0.523 & 0.506 & 0.495 & 0.473 & 0.526 \\
        Student-Request        & 0.000 & 0.038 & 0.029 & 0.087 & 0.125 & \textbf{0.200} & \textbf{0.200} & 0.000 & 0.033 & \textbf{0.200} \\
        Closed-Ended           & 0.077 & 0.095 & 0.118 & 0.262 & \textbf{0.657} & 0.629 & 0.644 & 0.650 & 0.623 & 0.635 \\
        Open-Ended             & 0.161 & 0.195 & 0.167 & 0.456 & \textbf{0.575} & 0.600 & 0.562 & 0.572 & 0.563 & 0.572 \\
        Task Related Prompt    & 0.012 & 0.006 & 0.065 & 0.117 & 0.000 & 0.222 & 0.194 & 0.136 & 0.040 & \textbf{0.344} \\
        \hline
        \textbf{Macro F1}      & 0.060 & 0.091 & 0.066 & 0.265 & 0.422 & 0.441 & 0.434 & 0.428 & 0.403 & \textbf{0.460} \\
        \hline
    \end{tabular}

    \caption{Per-label F1 scores of Instructional Discourse Labels for different large language models, prompted with zero-shot and few-shot prompts and fine-tuned model with different methods to overcome class imbalance: Binary Cross-Entropy loss (BCE), Focal loss (Focal), and Asymmetric loss (Asymmetric) in their normal variants (-), as well as their variants with positive label weighting (PW).}
    \label{tab:f1perlabeldisc}
\end{table*}

\end{document}